\pdfoutput=1

\documentclass[11pt]{article}

\usepackage[final]{acl}
\usepackage{times}
\usepackage{latexsym}
\usepackage{hyperref}
\usepackage{array}

\usepackage[T1]{fontenc}
\usepackage[utf8]{inputenc}
\usepackage{microtype}
\usepackage{inconsolata}
\usepackage{graphicx}
\usepackage{float}
\usepackage{listings}

\title{SplaXBERT: Leveraging Mixed Precision Training and Context Splitting for Question Answering}

\author{
    Zhu Yufan\textsuperscript{*}, Hao Zeyu\textsuperscript{*}, Li Siqi\textsuperscript{*}, Niu Boqian\textsuperscript{*} \\
    National University of Singapore, School of Computing \\
    \textsuperscript{*}Equal Contribution
}

\begin{document}
\maketitle

\begin{abstract}
SplaXBERT, built on ALBERT-xlarge with context-splitting and mixed precision training, achieves high efficiency in question-answering tasks on lengthy texts. Tested on SQuAD v1.1, it attains an Exact Match of 85.95\% and an F1 Score of 92.97\%, outperforming traditional BERT-based models in both accuracy and resource efficiency.
\end{abstract}

\section{Introduction} 
This project introduces SplaXBERT, a question-answering (QA) system optimized for handling lengthy, complex text passages. Built on the ALBERT-xlarge model, SplaXBERT is fine-tuned on the SQuAD v1.1 train set with mixed precision and evaluated on SQuAD v1.1 dev set. The model incorporates a unique context-splitting technique, segmenting long contexts into manageable chunks to enhance answer retrieval without additional model training, increasing both efficiency and adaptability.

Comparative testing with other BERT-based and emerging LLaMA models, along with ablation studies removing mixed precision and context splitting, highlights SplaXBERT's strengths. Results show that this combined strategy reduces information loss, improving answer precision and positioning SplaXBERT as a high-performance, resource-efficient QA solution for complex text scenarios.

\section{Preliminaries}

\subsection{ALBERT}

ALBERT (A Lite BERT) is a variant of BERT designed to reduce memory consumption and increase training speed without sacrificing performance. It uses two main techniques: factorized embedding parameterization and cross-layer parameter sharing. The factorized embedding parameterization decomposes the large vocabulary embedding matrix into two smaller matrices, significantly reducing parameters. Cross-layer parameter sharing shares parameters across layers, further decreasing the total number of parameters without compromising depth. These innovations allow ALBERT to achieve competitive results on natural language understanding tasks with fewer resources \cite{lan2020albert}.

\subsection{Mixed Precision FineTune}

Mixed precision training combines half-precision (FP16) and single-precision (FP32) computations to improve efficiency while preserving accuracy. Model weights, activations, and gradients are stored and processed in FP16, reducing memory usage and computational demands on GPUs. To counteract FP16’s narrower range, a master copy of weights is maintained in FP32, ensuring precise gradient updates. Loss scaling is applied to prevent small gradient values from becoming zero during backpropagation: scaling the loss up before computing gradients and then unscaling it before updating weights preserves essential gradient information. These techniques maintain model accuracy comparable to FP32-only training while significantly reducing memory consumption and increasing computation speed. The FP16 format might even add a regularization effect and improve prediction accuracy \cite{micikevicius2018mixedprecisiontraining}.

\subsection{Context Splitting}

Question-answering models like ALBERT face input size limitations with lengthy contexts. Context splitting mitigates this by dividing long passages into overlapping segments that fit within the model's maximum input length. Given a context \( C \) of \( N \) tokens and a maximum input length \( L \), we split \( C \) into \( k \) overlapping chunks \( \{C_1, C_2, \ldots, C_k\} \), each of length \( L \). The stride \( S \) determines the overlap, defined as \( S = L - O \), where \( O \) is the overlap size. The number of chunks \( k \) is calculated by:

\[
k = \left\lceil \frac{N - O}{L - O} \right\rceil
\]

Each segment is independently processed, and the highest-scoring answer is selected as the final output \cite{devlin2019bert}.

\section{Motivation: Challenges with LLaMA}
\label{llama}
First, we tested it on one of the most prevalent applications of large language models (LLMs) today, ChatBot-based QA. Intuitively, a ChatBot model appears well-suited for this task: it is inherently designed for question-answering, possesses a degree of logical reasoning, and offers flexibility for adaptation through prompting alone, without the need for additional retraining or fine-tuning.

We utilized the Llama 3 model, with 8B parameters, which, as of April 2024, is considered one of the most capable openly available LLMs \cite{llama3modelcard}. To assess its effectiveness in extractive question-answering tasks, we evaluated the model under two distinct experimental settings.

First, in the direct prompting approach, we provided the model with task instructions, response format, relevant context, and the question directly. This configuration aimed to test the model's ability to extract pertinent answers without additional prompt structuring or refinement.

In contrast, the prompt chaining with few-shot prompting approach employed a sequence of prompts using a few-shot learning technique. Specifically, the first LLM in the chain extracted relevant sentences from the provided context, while the second LLM refined these sentences by removing extraneous information, focusing solely on the precise information needed for the answer.

The experimental results are presented in Table \ref{tab:chatbot_result}. The exact match score reached approximately 70\%, while the F1 score was around 80\%. These metrics reflect a decent level of accuracy, indicating that the ChatBot model demonstrates a strong capability in handling extractive question-answering tasks without further finetuning.

\begin{table}[H]
    \centering
    \small
    \setlength{\tabcolsep}{2pt} 
    \begin{tabular}{l@{}c@{}c@{}}
        \hline
        Prompt Setting & Exact Match (\%) & F1 Score (\%) \\ 
        \hline
        Direct Prompting & 71.83 & 84.97 \\ 
        \hline
        Prompt Chaining with\\   Few-Shot Prompting & 71.13 & 80.93 \\
        \hline
    \end{tabular}
    \caption{Evaluation Results of Llama 3.2 ChatBot model}
    \label{tab:chatbot_result}
\end{table}

However, our findings suggest that implementing LLM chains did not enhance performance. Upon examining the generated responses, we observed instances where the output did not consistently adhere to the specified response format or other task requirements, resulting in suboptimal responses. Moreover, the use of two LLM chains appeared to increase the likelihood of such deviations, likely due to the compounded variability introduced by sequential processing. 

Despite the acceptable results, we identified several significant limitations in the ChatBot-based model during our experiments. First, the model often fails to retrieve the exact text from the context, likely due to the previously mentioned deviations in output formatting and structure. This inconsistency reduces its reliability in tasks requiring precise extraction. 

Second, the model's prediction latency presents a practical constraint. On average, it required approximately 8 seconds to answer a single extractive question, introducing a substantial delay. Such latency limits the model's feasibility for real-time or high-demand applications, where rapid response times are critical.

These limitations motivate us to explore alternative approaches, specifically through fine-tuning, to enhance the model's precision and efficiency in extractive question-answering tasks.

 \section{Related Works}
\label{sec:related_works}
The SQuAD dataset is widely recognized in question-answering research, with numerous studies evaluating model performance on it. Patel et al. compared six models, including XGBoost, Gaussian Mixture Models, and BERT, and found that BERT achieved the highest accuracy at 77\% \cite{patel2020comparativestudymachinelearning}. Devlin, the creator of BERT, emphasized in his paper that BERT's bidirectional structure enhances its suitability for fine-tuning on downstream tasks like SQuAD \cite{devlin2019bert}. Consequently, transformer-based models have become a natural choice for QA tasks. Özkurt compared various transformer models, finding that ALBERT outperformed models such as BERT-Medium, DistilBERT, and RoBERTa on the SQuAD v2.0 dataset \cite{Patel2020ComparativeSO}. Additionally, ALBERT is lighter than other BERT variants, as noted by the Hugging Face list of pre-trained models \cite{huggingface_transformers_2.4.0}, making it ideal for this project given limited access to high-powered GPUs. This motivated our focus on fine-tuning ALBERT.

\section{Methodology and Performance}

\subsection{Experiment Setup}
We conducted experiments on NVIDIA RTX 4090 GPU, fine-tuning the models on the SQuAD v1.1 training set, predicting with the SQuAD v1.1 development set, and evaluating on evaluation-v2.0.

\subsection{Baseline Model: BERT Base Uncased}

The initial model architecture used for fine-tuning is the BERT base uncased model with a question-answering head, which consists of two output layers that predict the start and end positions of the answer span within a given context. Fine-tuning adjusts both the layers of BERT and the question-answering head to specialize the model for extracting answer spans in question-context pairs.

To prepare the data, each question-context pair was tokenized using BERT’s tokenizer. Since context passages frequently exceed BERT’s maximum token limit of 384, a sliding window approach was applied to segment these long contexts into overlapping parts. For each segment, an attention mask and token type IDs were generated. The attention mask distinguishes between relevant and padding tokens within each input, and token type IDs help the model distinguish tokens from the question and context portions.

The answer span is mapped from character-level annotations in the context to token-level positions within each segment. The character-level start and end positions of the answer are identified and translated to their respective token-level positions within each tokenized segment. For segments where the answer span is out of bounds, the start and end positions are assigned to the [CLS] token, signaling that this particular segment does not contain the answer. The pipeline for our BERT Base QA model is illustrated below:

\begin{figure}[H]
    \centering
    \includegraphics[width=0.4\textwidth]{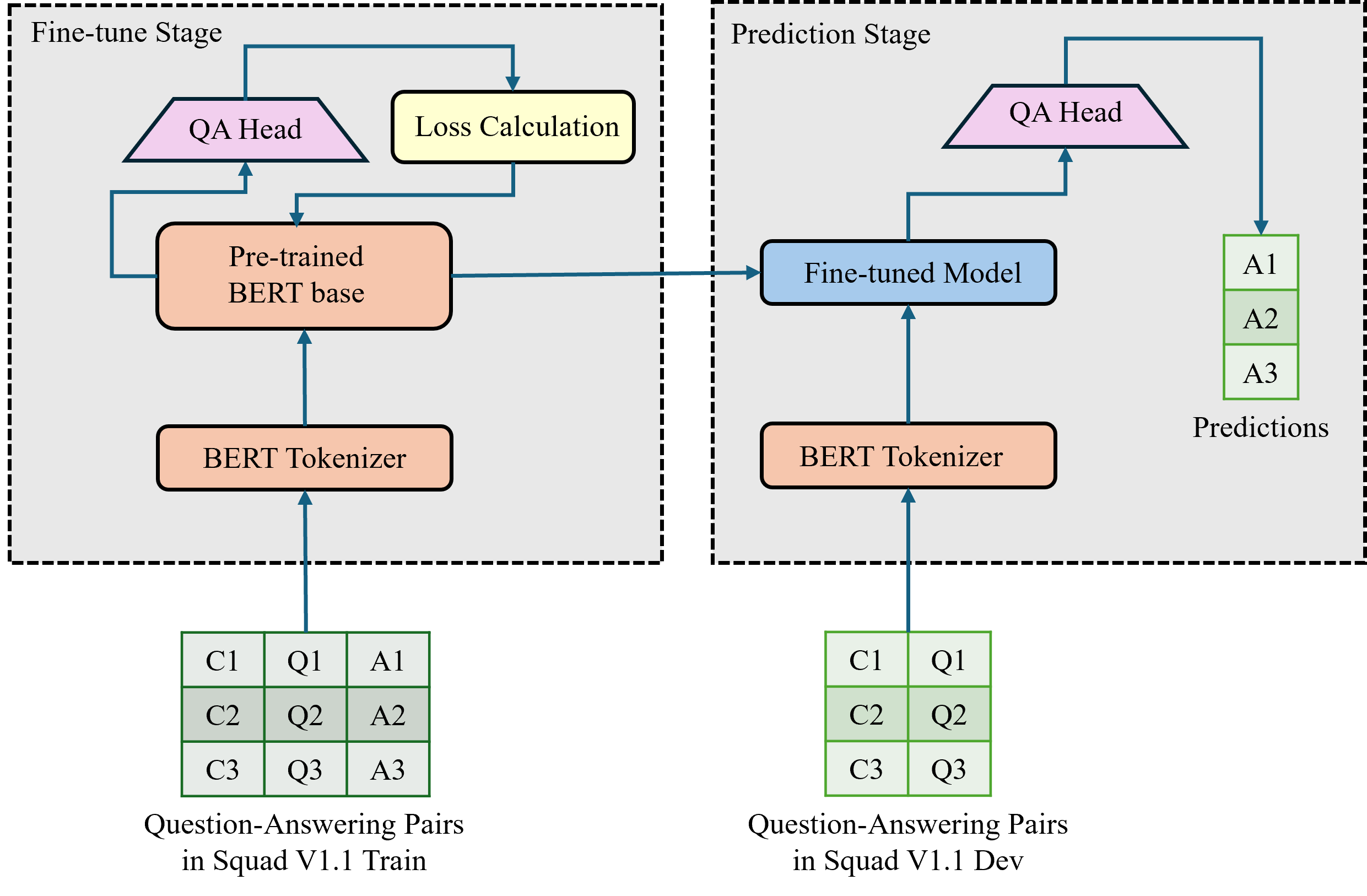}
    \caption{Overview of the BERT Base QA pipeline}
    \label{fig:model_architecture}
\end{figure}

The performance of the fine-tune model is shown below:

\begin{table}[h!]
    \centering
    \small
    \setlength{\tabcolsep}{2pt} 
    \begin{tabular}{l@{}c@{}c@{}}
        \hline
        Model & Exact Match (\%) & F1 Score (\%) \\ 
        \hline
        BERT Base \\
        Uncased (Fine-tuned) & 79.39 & 87.27 \\ 
        \hline
    \end{tabular}
    \caption{Evaluation Results of BERT Base Uncased Model on SQuAD v1.0}
    \label{tab:model_results}
\end{table}

This fine-tuned BERT Base Uncased model provides a solid baseline for our subsequent studies, serving as a reference point for evaluating the effectiveness of advanced techniques.

\subsection{Model Selection: ALBERT-xlarge}

While BERT Base Uncased with fine-tuning has achieved considerable success, as mentioned in the section \ref{sec:related_works}, we selected the pre-trained ALBERT-xlarge Uncased model due to its balance of model size and training efficiency. We keep the layers and techniques used in Bert Base but change the pre-trained model to ALBERT-xlarge. BERT Base contains 110 million parameters, whereas ALBERT Base reduces this to 11 million, and ALBERT-xlarge comprises 60 million parameters \cite{lan2020albert}. The reduced parameter count in ALBERT models leads to lower memory consumption and faster fine-tuning times compared to BERT. Given the time constraints of this project and our limited computational resources, ALBERT-xlarge strikes a balance between model complexity and computational efficiency.

\subsection{Mixed Precision Training}

We implemented mixed precision training packages from Pytorch. Specifically, auto-cast is used in the forward pass to automatically change the format from FP32 to FP16. Meanwhile, to prevent small gradients from going to zero, we used GradScaler as the scale to scale the loss up before doing backward propagation. In our empirical result, we trained ALBERT-xlarge on SQuAD v1.1 for three epochs. 

\begin{table}[h!]
    \centering
    \small
    \setlength{\tabcolsep}{2pt} 
    \begin{tabular}{l@{}c@{}c@{}}
        \hline
        Techniques & Exact Match (\%) & F1 Score (\%) \\ 
        \hline
        Without Mixed Precision & 85.32 & 91.75 \\ 
        \hline
        With Mixed Precision & 85.80 & 92.31 \\
        \hline
    \end{tabular}
    \caption{Evaluation Results of Fine-tuned Models on SQuAD v1.1}
    \label{tab:model_results}
\end{table}

In addition to enhancing accuracy, mixed precision training also reduced the training time per epoch by approximately 40 minutes, highlighting a significant improvement in efficiency.

\subsection{Context-Splitting Optimization}
\begin{figure}[H]
    \centering
    \includegraphics[width=0.45\textwidth]{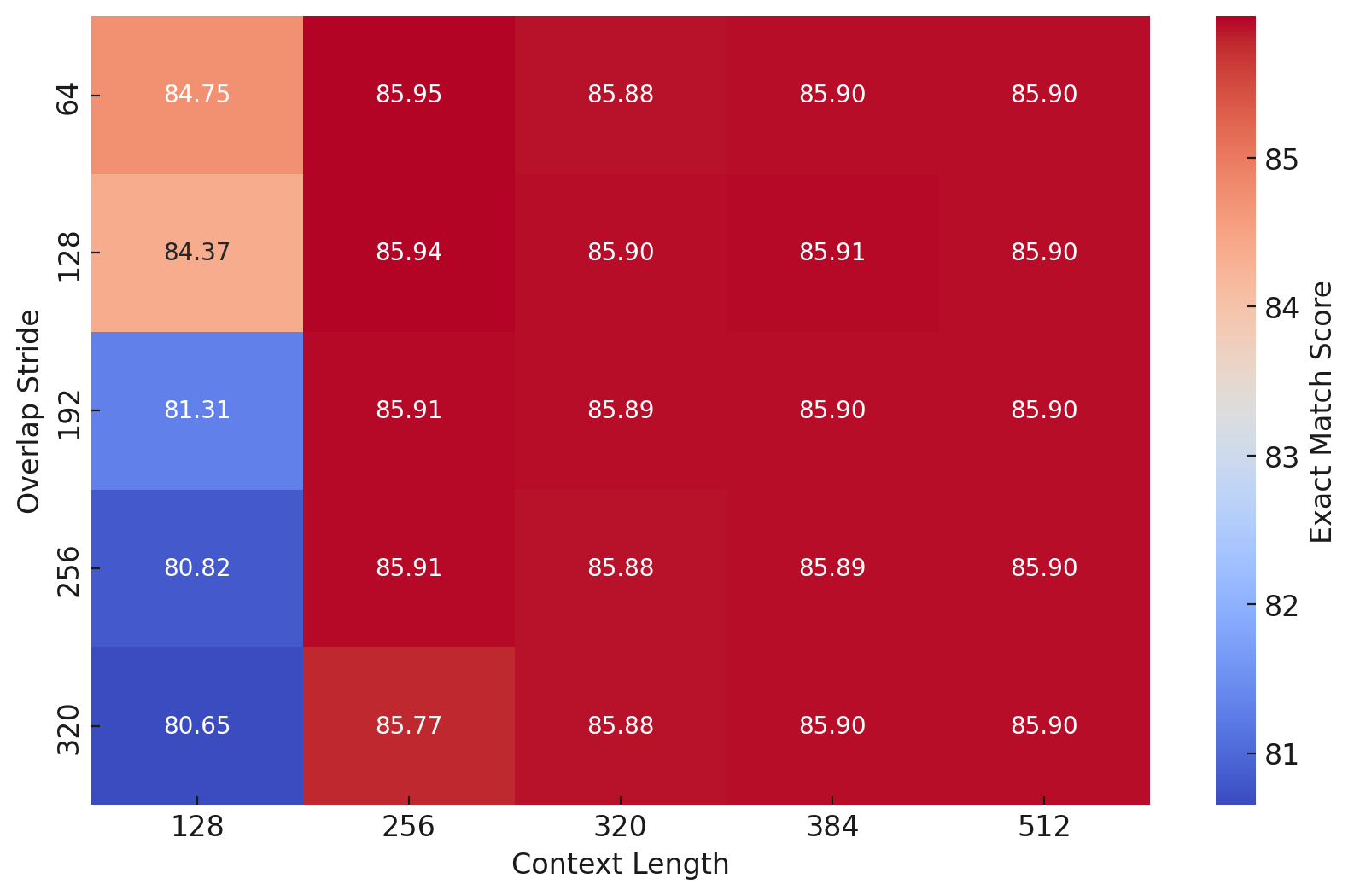}
    \caption{Grid Search Results: Exact Match Score by Context Length and Overlap Stride}
    \label{fig:exact_match}
\end{figure}

\begin{figure}[H]
    \centering
    \includegraphics[width=0.45\textwidth]{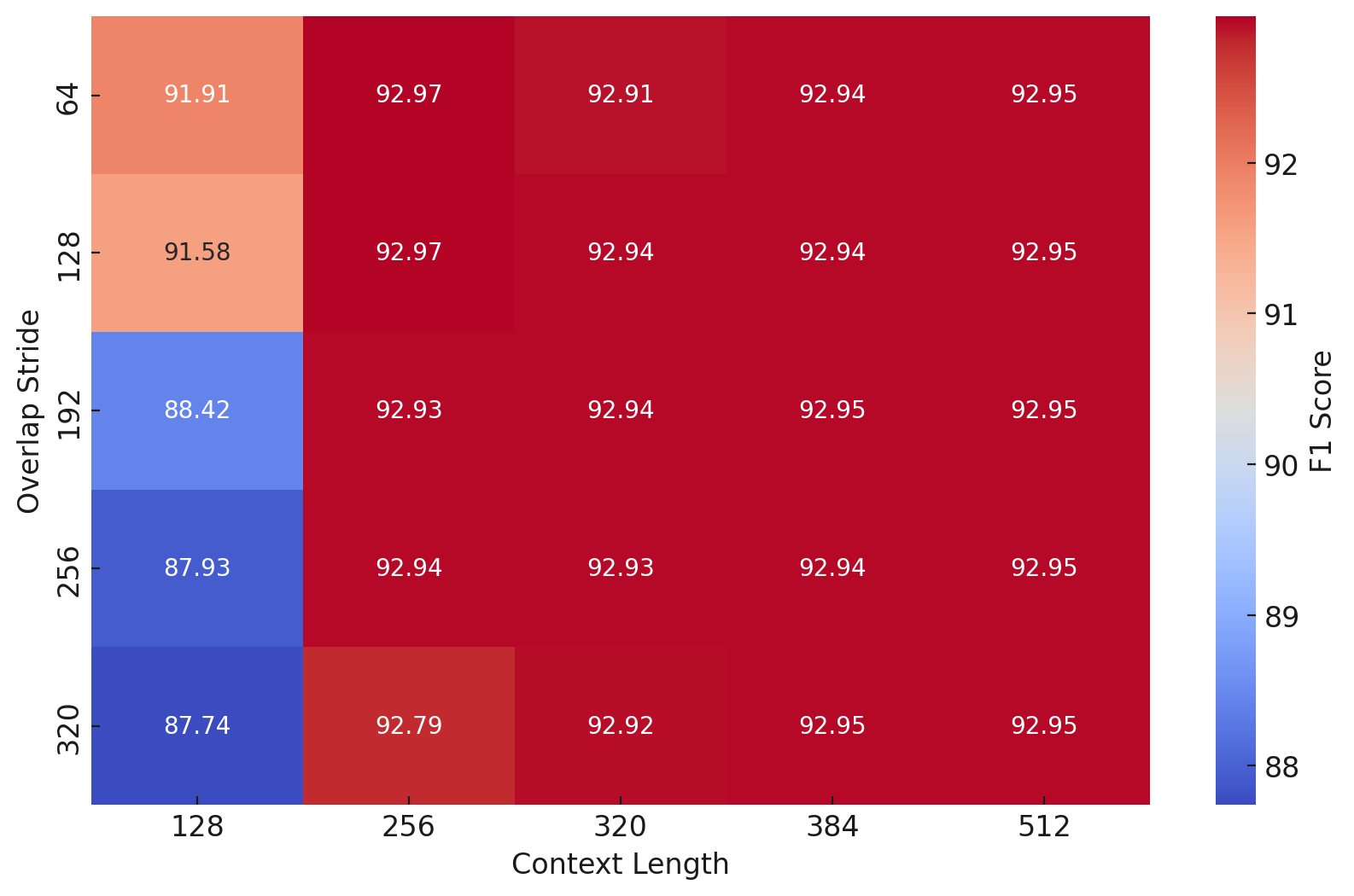}
    \caption{Grid Search Results: F1 Score by Context Length and Overlap Stride}
    \label{fig:f1_score}
\end{figure}

To improve accuracy during inference, we implemented a context-splitting technique that divides extended contexts into manageable segments. Given that inference is less time-intensive than training, we conducted a grid search to adjust segment length and overlap parameters for optimal answer retrieval accuracy. The grid search explored various combinations of segment length and overlap values, and the results, displayed in Figures~\ref{fig:exact_match} and~\ref{fig:f1_score}, show performance metrics for each configuration, specifically the Exact Match and F1 scores. These heatmaps suggest that a segment length of 128 to 256 and an overlap of 64 provide strong performance in both metrics, effectively balancing context length and overlap. This configuration maintains context continuity while minimizing information loss, thus enhancing the model’s ability to retrieve accurate answers without additional computational overhead.

\subsection{The Full Model Architecture}
Overall, our SplaXBERT model uses ALBERT-xlarge as the foundation for question-answering. We initialize the model with pre-trained weights and then apply mixed precision training on the SQuAD v1.1 development set for improved efficiency. During inference, we implement optimized context-splitting parameters to handle extended passages, maximizing answer retrieval accuracy while minimizing computational load. The pipeline is illustrated in the diagram below:

\begin{figure}[H]
    \centering
    \includegraphics[width=0.5\textwidth]{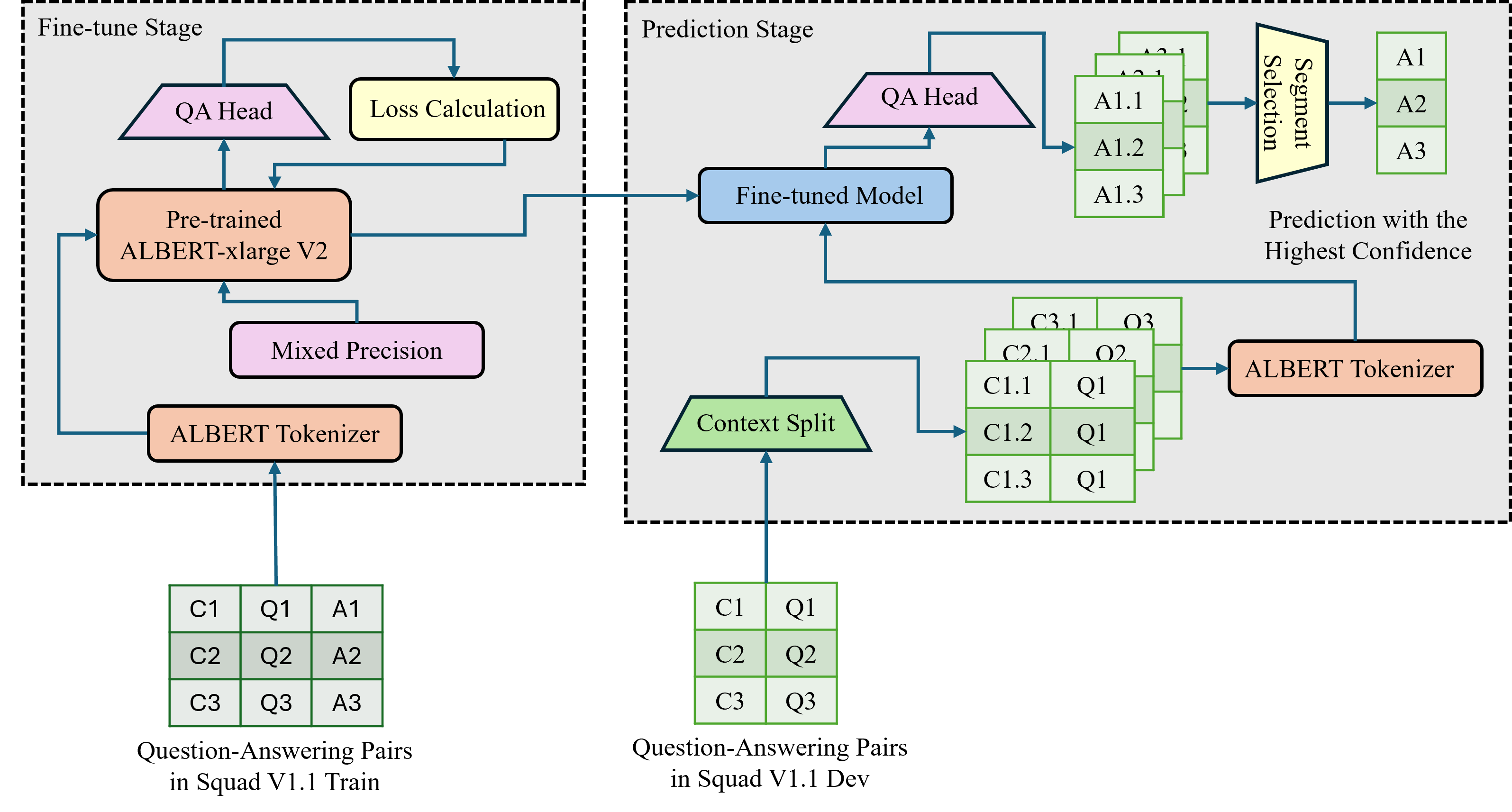}
    \caption{Overview of the SplaXBERT QA pipeline}
    \label{fig:model_architecture}
\end{figure}

\subsection{Summary of Performances}

The evaluation results of various fine-tuned models are presented in the table below. The results indicate that SplaXBERT achieves the highest Exact Match and F1 Score among the compared models.

\begin{table}[H]
    \centering
    \small
    \setlength{\tabcolsep}{2pt} 
    \begin{tabular}{l@{}c@{}c@{}}
        \hline
        Model & Exact Match (\%) & F1 Score (\%) \\ 
        \hline
        BERT Base (Fine-tuned) & 79.67 & 87.36 \\ 
        \hline
        BERT Distil Base (Fine-tuned) & 76.76 & 85.22 \\ 
        \hline
        BERT Base (Distilled \\   from BERT Large) & 77.57 & 85.77 \\
        \hline
        BERT Large (Fine-tuned) & 80.31 & 88.03 \\ 
        \hline
        ALBERT-base (Fine-tuned) & 75.62 & 84.61 \\ 
        \hline
        ALBERT-xLarge (Fine-tuned) & 85.32 & 91.75 \\
        \hline
        ALBERT-xLarge (Fine-tuned \\
        with Mixed Precision) & 85.80 & 92.31 \\
        \hline 
        SplaXBERT & 85.95 & 92.97 \\
        \hline 
    \end{tabular}
    \caption{Evaluation Results of Fine-tuned Models on SQuAD v1.1}
    \label{tab:summary_model_results}
\end{table}

\section{Conclusion}
In conclusion, the SplaXBERT project demonstrates the effectiveness of context-splitting and mixed precision training in enhancing question-answering accuracy and efficiency. By building on the ALBERT-xlarge model and segmenting lengthy passages, SplaXBERT achieves high answer retrieval accuracy without added model complexity. Despite limitations in computational resources, SplaXBERT’s performance highlights its potential as a resource-efficient solution for extractive QA tasks. Future work could leverage additional computational power to optimize performance further.

\newpage

\bibliography{cite}

\newpage
\indent
\newpage
\appendix
\section{Acknowledgments}
We express our sincere gratitude to Professor Ng Hwee Tou for his invaluable guidance and to our Teaching Assistant for his support.

The models ALBERT-base-v2, ALBERT-xlarge-v2, and BERT-base utilized in this work are pre-trained from external sources. Additionally, the integration of highway layers \cite{srivastava2015highwaynetworks} in the alternative studies draws inspiration from the work of the Stanford project group \cite{li2021ensemblealbertsquad20}. The Llama3 model is also pre-trained by external researchers \cite{llama3modelcard}. Our primary contributions include the fine-tuning of BERT and ALBERT models, the prompting of Llama3, and the development of a context-splitting methodology for improved prediction accuracy.

\section{Limitations}

This section presents the limitations of our work.

\subsection{GPU Power Constraints}
A major limitation of this project was the substantial GPU power required to fine-tune larger models. While we achieved promising results with the ALBERT Xlarge model, we could not explore more powerful variants, such as BERT XXLARGE. Fine-tuning these larger models would require thousands of GPU hours per epoch, making them impractical within the scope of this project. Consequently, we were unable to assess whether larger models might yield further performance improvements.

\subsection{Constraints on Hyperparameter Tuning}
The limited computational resources also restricted our ability to perform an exhaustive grid search for hyperparameter tuning. Although we conducted a grid search for optimal context-splitting parameters during inference, CPU limitations prevented us from exploring a broader range of configurations. This constraint likely impacted our ability to fine-tune some parameters fully, potentially affecting the model’s final performance.

\subsection{Context-Splitting in Inference Only}
Our use of context-splitting was limited to the inference stage. While integrating context-splitting during training could potentially yield additional improvements, we did not pursue this due to computational constraints and time limitations.

\subsection{Time Constraints}
The limited duration of this project further restricted our capacity to experiment with alternative layers and architectures. Exploring different model configurations could have enhanced performance, but this was beyond the scope given our project timeline and available resources.

\section{Alternative Studies}
\label{sec:appendix}
This section outlines alternative adjustments we explored, including the addition of Highway Layers and experiments with varying model depths.

\subsection{Adding Highway Layers}

The highway layer is a neural network architecture designed to facilitate training in very deep networks by allowing "information highways" through learned gating mechanisms \cite{srivastava2015highwaynetworks}. In a highway layer, the output is a combination of transformed input and a carry-over of the original input, regulated by two gates: the transform gate and the carry gate. The transform gate controls how much of the transformed input contributes to the output, while the carry gate governs how much of the original input is retained without modification. This structure is mathematically defined by \( y = H(x) \cdot T(x) + x \cdot (1 - T(x)) \), where \( H(x) \) is the transformation function, \( T(x) \) is the transform gate output, and \( x \) is the original input. By enabling some pathways to pass information with minimal transformation, highway layers reduce the vanishing gradient problem, making them suitable for training deep networks up to hundreds of layers.

Inspired by Li, Li \& Peng's effort to include highway layers in the fine-tuning of ALBERT \cite{li2021ensemblealbertsquad20}, we decided to experiment with the same thing on BERT. We added a highway layer immediately after the BERT model to form an ensemble model. The evaluation result of this model is shown below:

\begin{table}[h!]
    \centering
    \small
    \setlength{\tabcolsep}{2pt} 
    \begin{tabular}{l@{}c@{}c@{}}
        \hline
        Techniques & Exact Match (\%) & F1 Score (\%) \\ 
        \hline
        With Highway Layer & 79.29 & 87.09 \\ 
        \hline
        Without Highway Layer & 79.67 & 87.36 \\
        \hline
    \end{tabular}
    \caption{Evaluation Results of Fine-tuned Models on SQuAD v1.1}
    \label{tab:model_results}
\end{table}

Due to the disappointing result, we decided to stop using this ensemble model.

\subsection{Dataset Augmentation}

We experimented with training on SQuAD v2.0 while performing inference on SQuAD v1.1, anticipating improved performance based on preliminary trials with ALBERT-base. These initial tests showed promising improvements, as summarized in Table~\ref{tab:model_results_1}, where ALBERT-base fine-tuned on SQuAD v2.0 achieved higher Exact Match and F1 scores compared to fine-tuning on SQuAD v1.1 alone.

\begin{table}[H]
    \centering
    \small
    \setlength{\tabcolsep}{2pt} 
    \begin{tabular}{l@{}c@{}c@{}}
        \hline
        Techniques & Exact Match (\%) & F1 Score (\%) \\ 
        \hline
        Fine-tuned on v1.1 & 75.62 & 84.61 \\ 
        \hline
        Fine-tuned on v2.0 & 79.26 & 89.23 \\
        \hline
    \end{tabular}
    \caption{Evaluation Results of Fine-tuned ALBERT-base Models on SQuAD v2.0 with Inference on SQuAD v1.1}
    \label{tab:model_results_1}
\end{table}

However, this approach did not yield the same improvements with ALBERT-Xlarge, as shown in Table~\ref{tab:model_results_2}. ALBERT-Xlarge fine-tuned on both SQuAD v1.1 and v2.0 achieved nearly identical performance, suggesting that dataset augmentation with SQuAD v2.0 provided minimal benefit for this larger model.

\begin{table}[h!]
    \centering
    \small
    \setlength{\tabcolsep}{2pt} 
    \begin{tabular}{l@{}c@{}c@{}}
        \hline
        Techniques & Exact Match (\%) & F1 Score (\%) \\ 
        \hline
        Fine-tuned on v1.1 & 85.32 & 91.75 \\ 
        \hline
        Fine-tuned on v2.0 & 85.31 & 91.75 \\
        \hline
    \end{tabular}
    \caption{Evaluation Results of Fine-tuned ALBERT-Xlarge Models on SQuAD v2.0 with Inference on SQuAD v1.1}
    \label{tab:model_results_2}
\end{table}

\section{Prompts Used in Section \ref{llama}}
For clarity, we present the exact prompts used in Section \ref{llama}. Each prompt is tailored to ensure precise, contextually accurate responses from the model. We implemented two primary approaches: Direct Prompting and Prompt Chaining with Few-Shot Prompting, described below.

\subsection{Direct Prompting}
This prompt instructs the model to extract the exact answer from the provided passage without additional explanations or modifications.

\begin{quote}
    "Given the following passage, extract the answer to the following question from the passage.\\
    You should only respond with the EXACT SAME text from part of the passage (no slice within a single word). Make sure your answer directly addresses the question.\\
    NO EXPLANATION NEEDED. DO NOT REPEAT the question or include unnecessary information.\\
    \textbf{Passage:} \{passage\}\\
    \textbf{Question:} \{question\}"
\end{quote}

\subsection{Prompt Chaining with Few-Shot Prompting}
In this two-step approach, the model first generates an answer with the Direct Prompting method and then refines it through a second prompt designed to compress the response while retaining essential information.

\subsubsection{LLM Chain 1}
This chain uses the same prompt as Direct Prompting, generating an initial answer from the passage.

\subsubsection{LLM Chain 2}
This prompt refines the initial answer by compressing it to include only the key information necessary to answer the question concisely.

\begin{quote}
    "Given the question and answer below, extract and compress the answer to include only the key information needed to directly answer the question.\\
    DO NOT modify the text or add details; only select the most relevant portions of the text.\\
    
    First, ensure the response is MEANINGFUL and directly addresses the question (avoid trivial statements like 'AAA is AAA' or 'AAA is short for AAA').\\
    
    Second, ensure the response is as concise as possible, following these guidelines:
    \begin{itemize}
        \item Keep it to only a few words, if possible.
        \item Directly answer the question without repeating the question statement or completing a sentence unnecessarily.
        \item Avoid slicing within a single word; keep words whole.
        \item Only include conjunctions, prepositions, and linking words if specifically asked (e.g., respond with '2016' instead of 'since 2016' if not specifically required).
        \item Retain articles only if they appear in the original text; do not add them yourself.
        \item For questions asking for specific information (e.g., number, date, name, or location), respond with only that detail, omitting extra context (e.g., '2' instead of '2 apples').
        \item If asked for a team name, brand name, or other proper names, ensure the full name is intact without slicing.
    \end{itemize}
    
    Respond with ONLY the compressed answer in the EXACT SAME text, without modifications, extra words, or explanations.\\
    
    \textbf{Question:} \{question\}\\
    \textbf{Answer:} \{pred from LLM Chain 1\}"
\end{quote}

\section{Project Code and Sample Models for Reference}

The code used for this project is included in the submission, along with a comprehensive README file that explains the project structure and provides instructions for running our scripts. 

Throughout our exploration, we trained multiple models. However, due to file size constraints, we could not include all trained models in the submission. Instead, we have provided a selection of sample models as references. The scripts to train additional models and replicate our results are fully included in the submission.

\newpage
\onecolumn
\section*{Full-Size Figures}

For improved clarity, the model pipeline figures are presented here at full single-column width.

\begin{figure*}[h]
    \centering
    \includegraphics[width=0.7\textwidth]{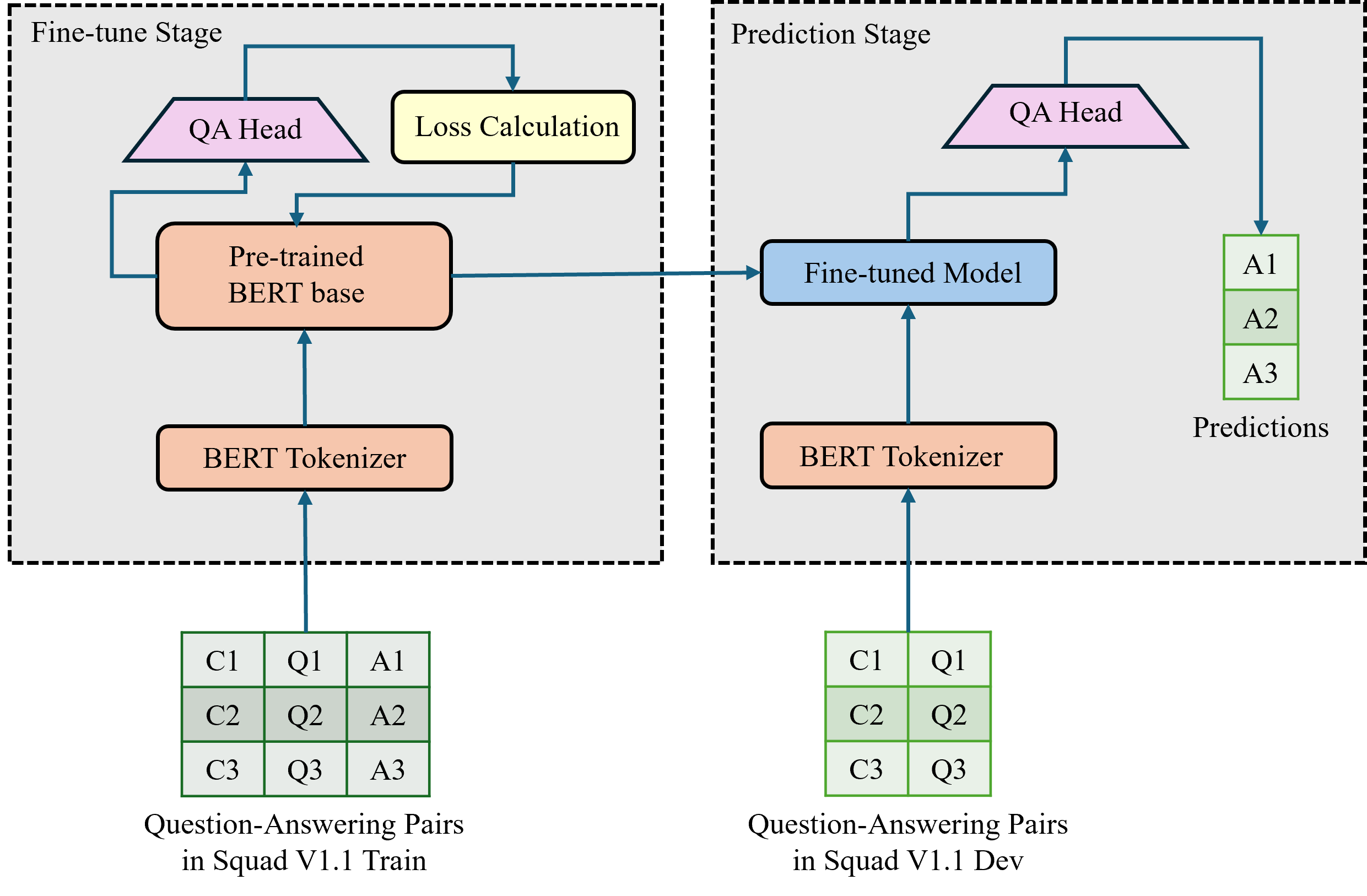}
    \caption{Overview of the BERT Base fine-tuning QA pipeline, showing the flow of data from input processing to answer extraction.}
    \label{fig:bert_architecture}
\end{figure*}

\begin{figure*}[h]
    \centering
    \includegraphics[width=0.9\textwidth]{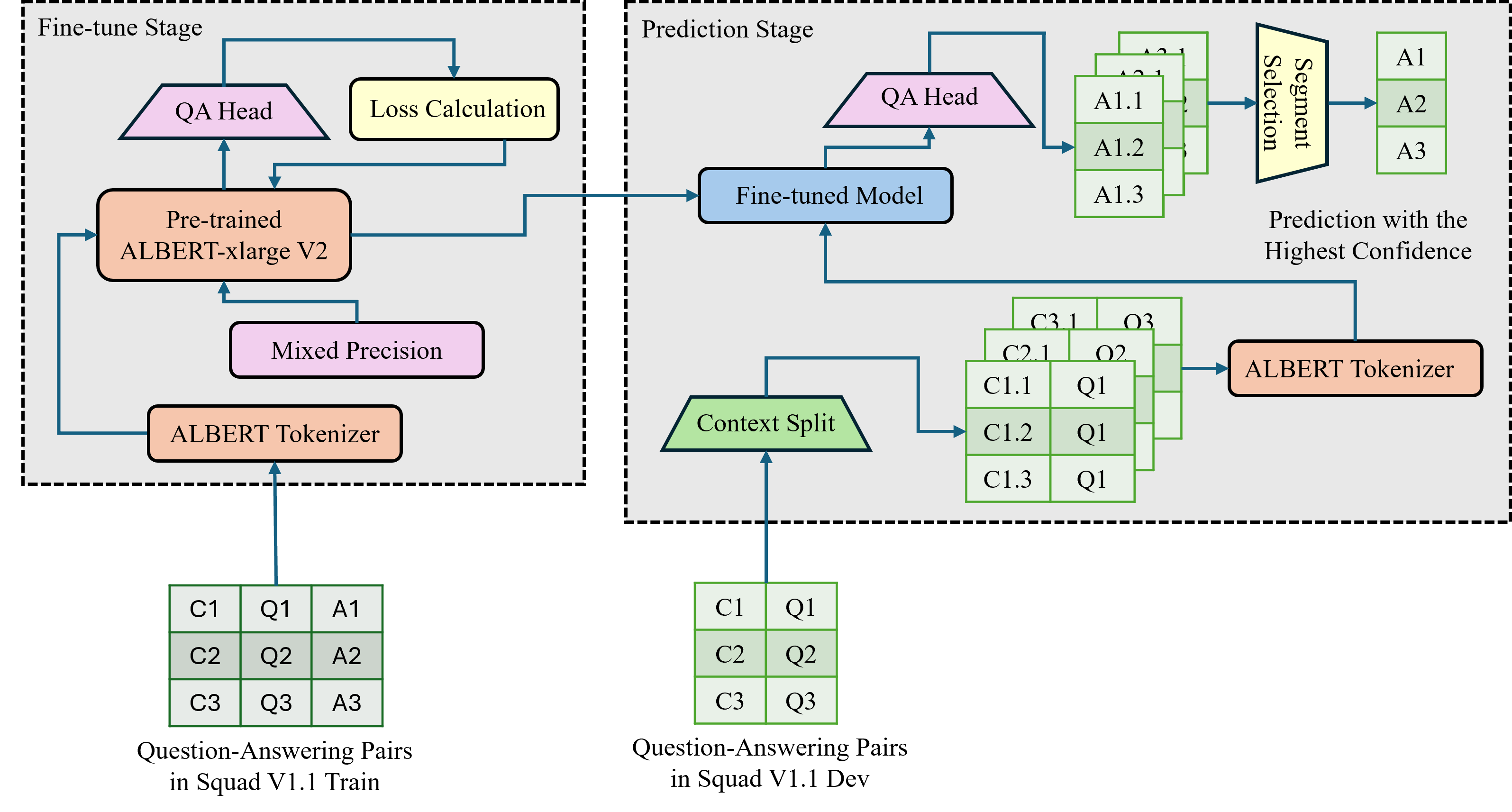}
    \caption{Overview of the SplaXBERT QA pipeline, highlighting the integration of context-splitting and mixed precision training for efficient answer retrieval.}
    \label{fig:splaxbert_architecture}
\end{figure*}

\end{document}